\documentclass[11pt,a4paper]{article}
\usepackage[hyperref]{acl2021}
\usepackage{times}
\usepackage{latexsym}
\usepackage{graphicx}
\usepackage{amsmath}
\usepackage{booktabs}
\usepackage{multirow}
\usepackage[frozencache,cachedir=.]{minted}

\newcommand*\samethanks[1][\value{footnote}]{\footnotemark[#1]}

\usepackage{microtype}

\aclfinalcopy 
\def\aclpaperid{20} 

\title{Goal-Oriented Script Construction}

\author{Qing Lyu\thanks{\hspace{4pt} Equal contribution.} \\\And
  Li Zhang\samethanks \\
  University of Pennsylvania \\
  {\tt \{lyuqing,zharry,ccb\}@seas.upenn.edu} \\\And
  Chris Callison-Burch \\}

\date{}

\begin{document}
\maketitle
\begin{abstract}
The knowledge of scripts, common chains of events in stereotypical scenarios, is a valuable asset for task-oriented natural language understanding systems. We propose the Goal-Oriented Script Construction task, where a model produces a sequence of steps to accomplish a given goal. We pilot our task on the first multilingual script learning dataset supporting 18 languages collected from wikiHow, a website containing half a million how-to articles. For baselines, we consider both a generation-based approach using a language model and a retrieval-based approach by first retrieving the relevant steps from a large candidate pool and then ordering them. We show that our task is practical, feasible but challenging for state-of-the-art Transformer models, and that our methods can be readily deployed for various other datasets and domains with decent zero-shot performance\footnote{Our models and data are be available at \url{https://github.com/veronica320/wikihow-GOSC}.}.

\end{abstract}

\section{Introduction}

A \textit{script} is a standardized sequence of events about stereotypical activities \cite{feigenbaum1981handbook}. For example, ``\textit{go to a restaurant}'' typically involves ``\textit{order food}'', ``\textit{eat}'', ``\textit{pay the bill}'', etc. Such script knowledge has long been proposed as a way to enhance AI systems \cite{abelson1977scripts}. Specifically, task-oriented dialog agents may greatly benefit from the understanding of \textit{goal-oriented scripts}\footnote{\url{https://developer.amazon.com/alexaprize}}. However, the evaluation of script knowledge remains an open question \cite{chambers2017behind}. Moreover, it is unclear whether current models can generate complete scripts. Such an ability is in high demand for recent efforts to reason about complex events \cite{li2020connecting, wen-etal-2021-resin}\footnote{\href{https://tinyurl.com/yxwztj3j}{{\tt www.darpa.mil/program/knowledge- directed-artificial-intelligence-reasoning -over-schemas}}}. 

We propose the task of \textit{Goal-Oriented Script Construction} (GOSC) to \textit{holistically} evaluate a model's understanding of scripts. Given a \textit{goal} (or the name of a script), we ask the model to construct the sequence of \textit{steps} (or events in a script) to achieve the goal. This task targets a model's ability to narrate an entire script, subsuming most existing evaluation tasks. Our rationale is that a model that \textit{understands} some scripts (e.g. how to ``\textit{travel abroad}'' and ``\textit{go to college}'') should be able to produce \textit{new} ones (e.g. how to ``\textit{study abroad}'') using the absorbed knowledge, close to how humans learn.

\begin{figure}[t!]
    \centering
    \includegraphics[scale=0.175]{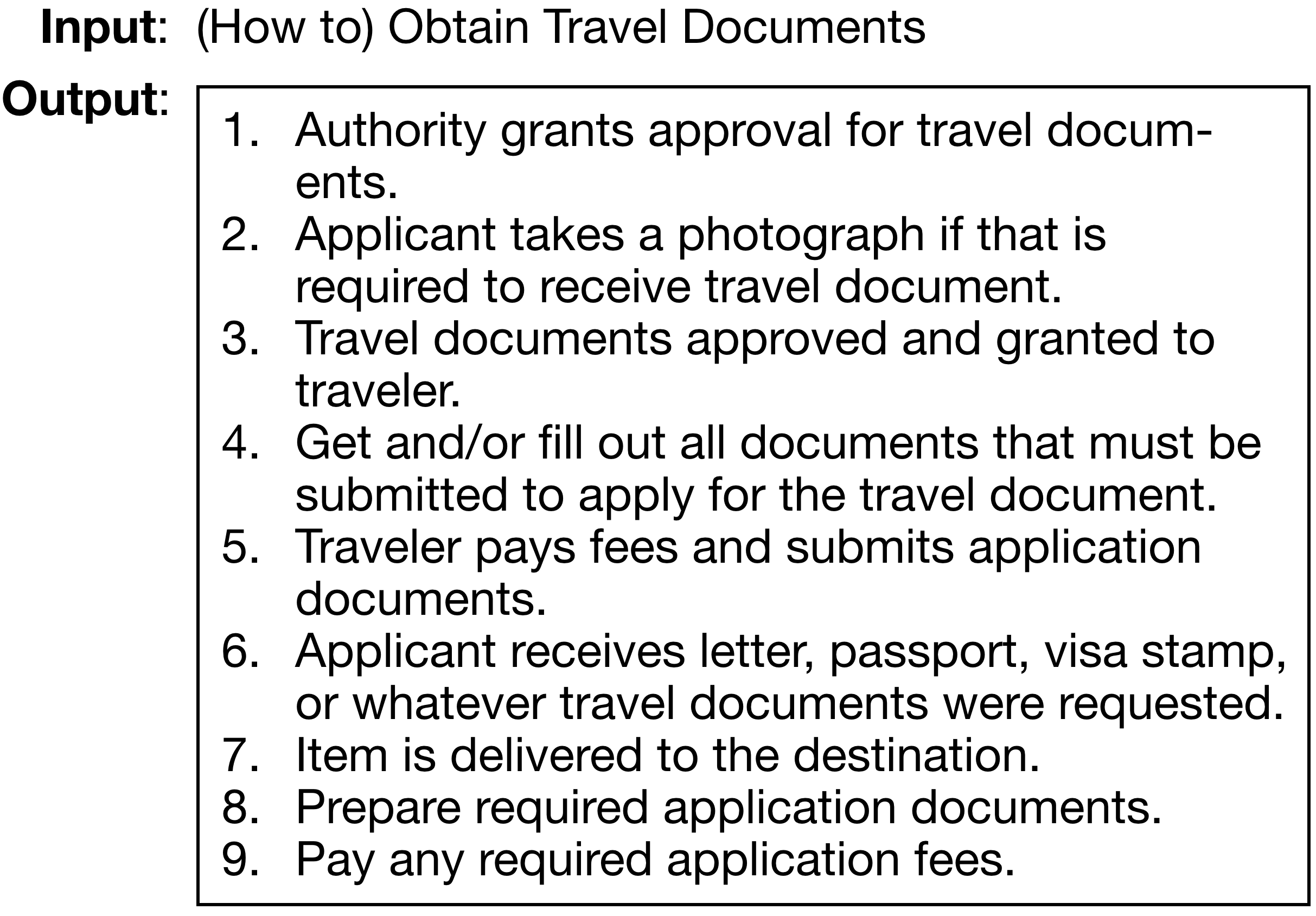}
    \caption{An example script constructed by our Step-Inference-Ordering pipeline in a zero-shot manner. The input is a \textit{goal}, and the output is an ordered list of steps.}
    \label{fig:thumbnail_script}
\end{figure}

While almost all prior script learning work has focused on English, we introduce a novel multilingual corpus. Our corpus is collected from wikiHow (\url{wikihow.com}), a website of how-to articles in 18 languages. The articles span a wide range of domains, from commonplace activities like going to a restaurant to more specific ones like protecting oneself from the coronavirus.  



We train and evaluate several baseline systems on our GOSC task. First, we consider a generation-based approach where a pretrained language model, multilingual T5, is finetuned to produce scripts from scratch. As an alternative, observing that most desired steps can be drawn from the training scripts due to their magnitude and high coverage, we also propose a retrieval-based approach. Concretely, we develop a Step-Inference-Ordering pipeline using existing models to retrieve relevant steps and order them. We also improve the pipeline with techniques such as multitask learning. From the experiments, the GOSC task proves challenging but feasible for state-of-the-art Transformers. Furthermore, we show that our pipeline trained on wikiHow can generalize to other datasets and domains (see an example in Figure \ref{fig:thumbnail_script}). On three classic script corpora, OMICS, SMILE, and DeScript, it achieves strong zero-shot performance. It can also be directly deployed to construct scripts in distant domains (e.g. military/political).

In this paper, we make several contributions:\\
1) We propose the GOSC task targeting the comprehensive understanding of scripts. \\
2) We introduce the first multilingual script learning dataset available in 18 languages.\\
3) We compare generation-based and retrieval-based approaches using both automatic and human judgments, which demonstrate the feasibility but also the difficulty of GOSC.\\
4) We show that our approach can be readily applied to other datasets or other domains.

\section{Related Work}

The notion of \textit{scripts} \cite{abelson1977scripts}, or \textit{schemas} \cite{rumelhart1975notes}, encodes the knowledge of standardized event sequences. We dissect previous work on script learning into two lines, \textit{narrative} and \textit{procedural}.

One line of work focuses on \textit{narrative} scripts, where \textit{declarative}, or \textit{descriptive} knowledge is distilled from narrative texts like news or stories \cite{mujtaba2019recent}. Such scripts are not goal-oriented, but descriptions of sequential events (e.g. a traffic accident involves a collision, injuries, police intervention, etc.). \citet{chambers2008unsupervised} introduced the classic Narrative Cloze Test, where a model is asked to fill in the blank given a script with one missing event. Following the task, a few papers made extensions on representation \cite{chambers2009unsupervised, pichotta2014statistical} or modeling \cite{jans2012skip, pichotta2016statistical,pichotta2016learning,pichotta2016using}, achieving better performance on Narrative Cloze. Meanwhile, other work re-formalized Narrative Cloze as language modeling (LM) \cite{rudinger2015script} or multiple-choice \cite{granroth2016happens} tasks. However, the evolving evaluation datasets contain more spurious scripts, with many uninformative events such as ``say'' or ``be'', and the LMs tend to capture such cues \cite{chambers2017behind}.

The other line of work focuses on \textit{procedural} scripts, where events happen in a scenario, usually in order to achieve a goal. For example, to ``visit a doctor'', one should ``make an appointment'', ``go to the hospital'', etc. To obtain data, Event Sequence Descriptions (ESD) are collected usually by crowdsourcing, and are cleaned to produce scripts. Thus, most such datasets are small-scale, including OMICS \cite{singh2002open}, SMILE \cite{regneri2010learning}, the \citet{li2012crowdsourcing} corpus, and DeScript \cite{wanzare2016crowdsourced}. The evaluation tasks are diverse, ranging from event clustering, event ordering \cite{regneri2010learning}, text-script alignment \cite{ostermann2017aligning} and next event prediction \cite{nguyen-etal-2017-sequence}. There are also efforts on domain extensions \cite{yagcioglu2018recipeqa, berant2014modeling} and modeling improvements \cite{frermann2014hierarchical, modi2014inducing}.

In both lines, it still remains an open problem what kind of automatic task most accurately evaluates a system's understanding of scripts. Most prior work has designed tasks focusing on various fragmented pieces of such understanding. For example, Narrative Cloze assesses a model's knowledge for completing a close-to-finished script. The ESD line of work, on the other hand, evaluates script learning systems with the aforementioned variety of tasks, each touching upon a specific piece of script knowledge nonetheless. Recent work has also brought forth generation-based tasks, but mostly within an open-ended/specialized domain like story or recipe generation \cite{fan-etal-2018-hierarchical, xu-etal-2020-megatron}. 

Regarding data source, wikiHow has been used in multiple NLP efforts, including knowledge base construction \cite{jung2010automatic, chu2017distilling}, household activity prediction \cite{nguyen-etal-2017-sequence}, summarization \cite{Koupaee2018WikiHowAL, ladhak-etal-2020-wikilingua}, event relation classification \cite{park2018learning}, and next passage completion \cite{zellers-etal-2019-hellaswag}. A few recent papers \cite{zhou-etal-2019-learning, zhang-etal-2020-reasoning} explored a set of separate goal-step inference tasks, mostly in binary-classification/multiple-choice formats, with few negative candidates. Our task is more holistic and realistic, simulating an open-ended scenario with retrieval/generation settings. We combine two of our existing modules from \citet{zhang-etal-2020-reasoning} into a baseline, but a successful GOSC system can certainly include other functionalities (e.g. paraphrase detection). Also similar is \citet{zhang-etal-2020-analogous}, which doesn't include an extrinsic evaluation on other datasets/domains though.

In summary, our work has the following important differences with previous papers:\\
1) Existing tasks mostly evaluate fragmented pieces of script knowledge, while GOSC is higher-level, targeting the ability to invent \textit{new, complete} scripts.\\
2) We are the first to study \textit{multilingual} script learning. We evaluate several baselines and make improvements with techniques like multitask learning.\\
3) Our dataset improves upon the previous ones in multiple ways, with higher quality than the mined narrative scripts, lower cost and larger scale than the crowdsourced ESDs.\\
4) The knowledge learned from our dataset allows models to construct scripts in other datasets/domains without training.

\section{Goal Oriented Script Construction} \label{section:task}

We propose the Goal-Oriented Script Construction (GOSC) task. Given a \textit{goal} $g$, a system constructs a complete script as an ordered list of \textit{steps} $S$, with a ground-truth reference $T$. As a hint of the desired level of granularity, we also provide an expected number of steps (or length of the script), $l$, as input. Depending on whether the set of possible candidate steps are given in advance, GOSC can happen in two settings: Generation or Retrieval. 

In the \textbf{Generation setting}, the model must generate the entire script from scratch. 

In the \textbf{Retrieval setting}, a large set of candidate steps $C$ is given. The model must predict a subset of steps $S$ from $C$, and provide their ordering. 

\vspace{-0.04in}
\section{Multilingual WikiHow Corpus} \label{section:corpus}
\vspace{-0.04in}

Our previously wikiHow corpus \cite{zhang-etal-2020-reasoning} is a collection of how-to articles in English (en). We extend this corpus by crawling wikiHow in 17 other languages, including Spanish (es), Portuguese (pt), Chinese (zh), German (de), French (fr), Russian (ru), Italian (it), Indonesian (id), Dutch (nl), Arabic (ar), Vietnamese (vn), Thai (th), Japanese (jp), Korean (ko), Czech (cz), Hindi (hi), and Turkish (tr). The resulting multilingual wikiHow corpus may be used in various tasks in NLP and other fields.

For script learning, we extract from each wikiHow article the following critical components to form a \textit{goal-oriented script}.\\
\textbf{Goal}: the title stripped of ``How to'';\\
\textbf{Section}: the header of a ``method'' or a ``part'' which contains multiple steps;\footnote{We ignore this hierarchical relation and flatten all steps in all Sections as the Steps of the script. }\\
\textbf{Steps}: the headlines of step paragraphs; \\
\textbf{Category}: the top-level wikiHow category. \\
An example wikiHow script is shown in Figure~\ref{json_example}.

\begin{figure}
\begin{minted}[linenos=false,
               fontsize=\small]{js}
{
 "title": "Eat at a Sit Down Restaurant",
 "category": "FOOD AND ENTERTAINING",
 "ordered": True,
 "sections": [ ...
  {
   "section": "Ordering Out",
   "steps": [ ...
    "Order drinks first.",
    "Ask about daily specials.",
    "Look over the menu and place your 
     food order.", ...
   ],
  }, ... ]}
\end{minted}
\caption{An abridged example script extracted from the English wikiHow article ``How to Eat at a Sit Down Restaurant''.} 
\label{json_example}
\end{figure}

Our previous corpus provides labels of whether each English article is ordered, predicted by a high-precision classifier. We project these labels to other languages using the cross-language links in each wikiHow article. For articles without a match to English, it defaults to unordered. In our task setup, we only require the model to order the steps if an article is ordered.

For all experiments below, we randomly hold out 10\% articles in each language as the test set, and use the remaining 90\% for training and development.\footnote{See Appendix~\ref{appendix:corpus_stats} for our corpus statistics.}

\begin{figure*}[t!]
    \centering
    \includegraphics[scale=0.47]{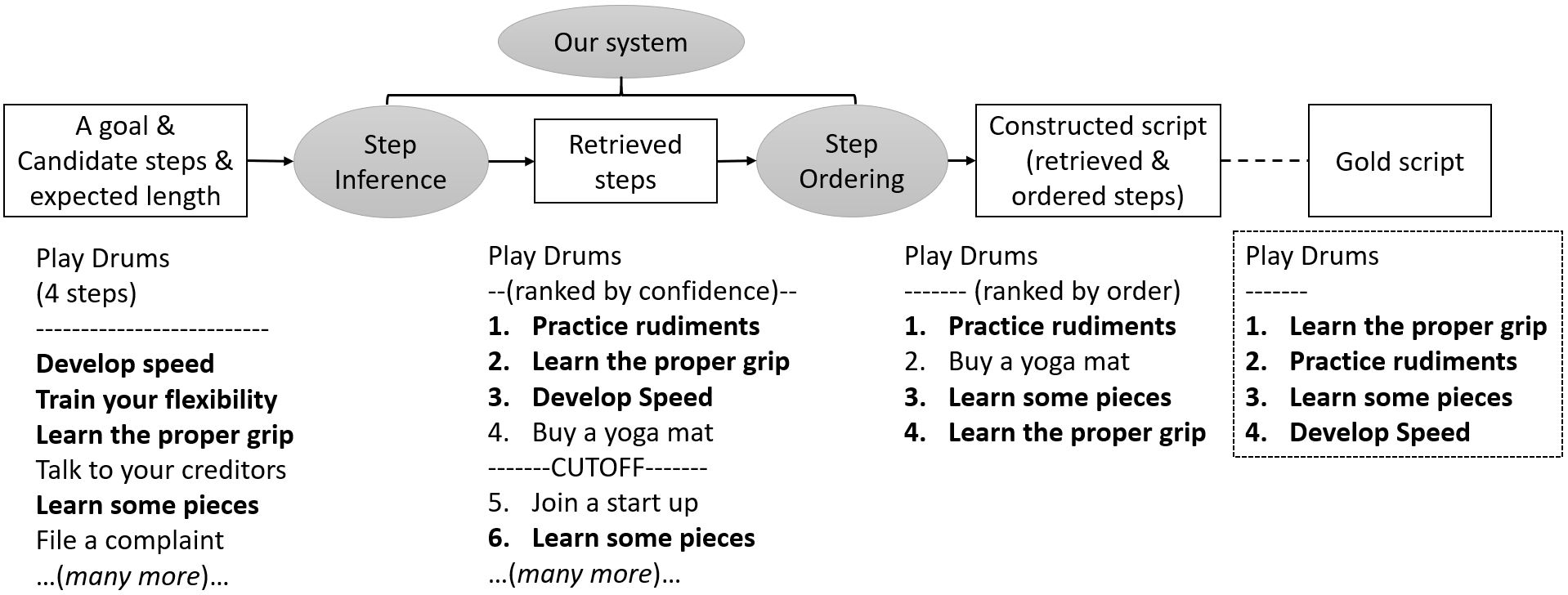}
    \caption{Our Step-Inference-Ordering pipeline for the GOSC Retrieval task. An example \textit{ordered} script is shown with example steps in the input and output. Those that appear in the ground-truth script is in bold.}
    \label{fig:pipeline}
\end{figure*}

We use the corpus to construct a dataset for multilingual GOSC. For the Retrieval setting, the set of candidate steps $C$ are all the steps present in the test set. However, we observe that not only the large number of steps may render the evaluation intractable, but most steps are also evidently distant from the given goal. To conserve computing power, we restrict $C$ as all the steps from articles within the same wikiHow category for each script.

\section{Models}

We develop two systems based on state-of-the-art Transformers for the GOSC task.\footnote{Reproducibility details can be found in Appendix~\ref{appendix:model}.} 

\subsection{Generation Approach: Multilingual T5}
\label{section:generation_baseline}

For the Generation setting, we finetune mT5 \cite{xue2021mt5}, a pretrained generation model that is not only state-of-the-art on many tasks but also the only available massively multilingual one to date. 

During finetuning, we provide the goal of each article in the training set as a prompt, and train the model to generate the sequence of all the steps conditioned on the goal. Therefore, the model's behavior is similar to completing the task of inferring relevant steps and sorting them at once. At inference time, the model generates a list of steps given a goal in the test set. 

\subsection{Retrieval Approach: Step-Inference-Ordering Pipeline}
\label{section:step_inf_order_pipeline}

We then implement a \textit{Step-Inference-Ordering pipeline} for the Retrieval setting. Our pipeline contains a Step Inference model to first gather the set of desired steps, and a Step Ordering model to order the steps in the set. These models are based on our previous work \cite{zhang-etal-2020-reasoning}. Under the hood, the models are pretrained XLM-RoBERTa \cite{conneau-etal-2020-unsupervised} or mBERT \cite{devlin-etal-2019-bert} for binary classification, both state-of-the-art multilingual representations.

Our Step Inference model takes a goal and a candidate step as input, and outputs whether the candidate is indeed a step toward the goal with a confidence score. During training, for every script, its goal forms a positive example along with each of its steps. We then randomly sample 50 steps from other scripts within the same wikiHow category and pair them with the goal as negative examples. The model predicts a label for each goal-step pair with a cross-entropy loss. During evaluation, for each script in the test set, every candidate step is paired with the given goal as the model input. We then rank all candidate steps based on the model confidence scores decreasingly. Finally, the top $l$ steps are retained, where $l$ is the required length. 

Our Step Ordering model takes a goal and two steps as input, and outputs which step happens first. During training, we sample every pair of steps in each ordered script as input to the model with a cross-entropy loss. During evaluation, we give every pair of retrieved steps as input, and count the total number of times that a step is ranked before others. We then sort all steps by this count to approximate their complete ordering.

\begin{figure*}
    \centering
    \includegraphics[scale=0.15]{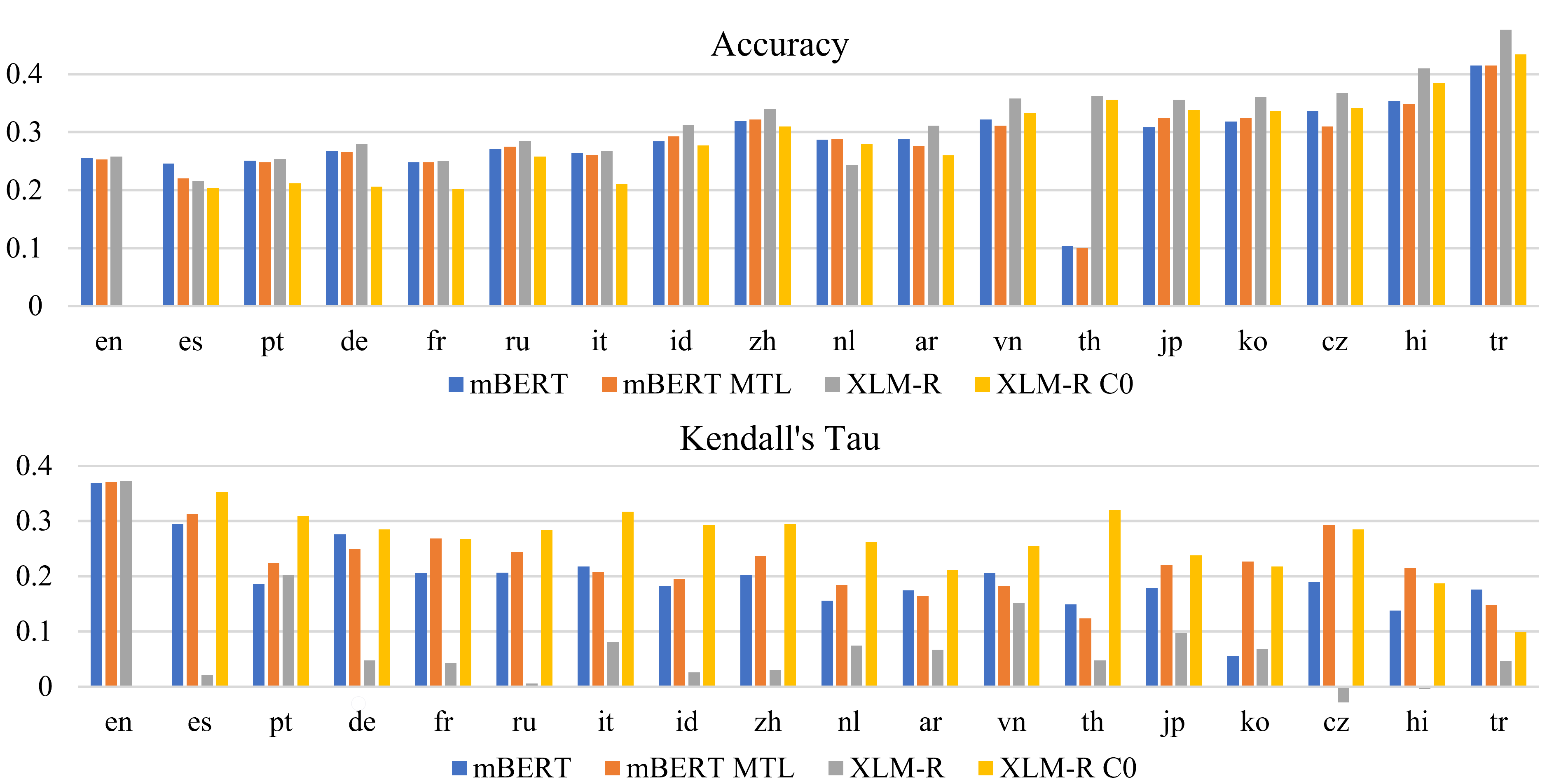}
    \caption{Detailed performance on each language from Table~\ref{table:retrieval_avg_results}.}
    \label{fig:charts}
\end{figure*}

An illustration of our Step-Inference-Ordering pipeline is shown in Figure~\ref{fig:pipeline}. We also consider two additional variations.\\
\textbf{Multitask Learning} (MTL): The Step Inference and the Step Ordering models share the encoder layer, but have separate classifier layers. During training, the MTL system is then presented with a batch of examples from each task in an alternating fashion. During evaluation, the corresponding classifier is used. \\
\textbf{Cross-Lingual Zero-Shot Transfer} (C0): While there are abundant English training scripts, data in some other languages are scarce. Hence, we also attempt to directly evaluate the English-trained models on non-English data. 

\section{In-Domain Evaluation}
\label{section:in_domain_eval}
To demonstrate the performance of models on the GOSC task, we evaluate them on our multilingual wikiHow dataset using both automatic metrics and human judgments. The ultimate utility for this task is the extent to which a human can follow the constructed steps to accomplish the given goal. As direct user studies might be costly and hard to standardize, we carefully choose measures that adhere to this utility. By default, all models are trained and evaluated on the same language.

\begin{table}
\small
\begin{tabular}{ p{0.55cm}|p{0.55cm}p{0.55cm}p{0.55cm}cccccc } 
\toprule
Lang. & \hspace{0.2em} en & es & pt & de & fr & ru  \\
\midrule
Perp. & \hspace{0.2em} 17 & 11 & 24 & 97 & 46 & 79  \\
Bert. &  .823 & .702 & .682 & .677 & .718 & .682  \\
\midrule
\midrule
Lang. & \hspace{0.2em} it & id & zh & nl & ar & vn \\
\midrule
Perp. & \hspace{0.05em} 116 & 269 & 13,249 & 955 & 746 & 97  \\
Bert. &  .653 & .692 & .667 & .690 & .701 & .695  \\
\midrule
\midrule
Lang. & \hspace{0.2em} th & jp & ko & cz & hi & tr \\
\midrule
Perp. & 29,538 & 73,952 & 2,357 & 1,823 & 2,033 & 36,848  \\
Bert. & .701 & .679 & .692 & .682 & .704 & .665  \\
\bottomrule
\end{tabular}
\caption{Auto evaluation results for the Generation setting (Perplexity and BERTScore F1 measure). The performance of multilingual T5 is reported.}
\label{table:generation_ppl}
\end{table}

\begin{table}
\small
\centering
\resizebox{7.7cm}{!}{%
\begin{tabular}{ l|cc|cccc|c|cc } 
 \toprule
\multirow{2}{1.6cm}{Model} & \multicolumn{2}{c|}{English only}  & \multicolumn{2}{c}{Avg. all lang.s} \\
& Acc. & Kendall's $\tau$  & Acc.   & Kendall's $\tau$   \\
 \midrule
mBERT & .256 & .369 & .286 & .198\\
mBERT MTL  & .253  & .371 & .283 & .226\\
XLM-R & .258  & .372 & .317 & .075\\
XLM-R C0  & -  & - & .291 & .264\\
\bottomrule
\end{tabular}%
}
\caption{Auto evaluation results for the Retrieval setting (Accuracy and Kendall's Tau). The performance of mBERT and XLM-RoBERTa, along with their multitask (MTL) and crosslingual zero-shot transfer (C0) variations, are reported \footnotemark.}
\label{table:retrieval_avg_results}
\end{table}
\footnotetext{Multitask XLM-R and cross-lingual zero-shot mBERT are found to perform a lot worse and thus omitted.}

\subsection{Auto Evaluation for Generation Setting}
\label{section:generation_results}
To automatically evaluate models in the Generation Setting, we report \textbf{perplexity} and \textbf{BERTScore} \cite{zhang2019bertscore}, as two frequently used metrics for evaluating text generation.

The mean perplexity of mT5 on the test set of each language is shown in Table~\ref{table:generation_ppl}. The results show a large range of variation. To see if perplexity correlates with the data size, we conduct a Spearman's rank correlation two-tailed test. We find a Spearman's $\rho$ of $-0.856$ and a p-value of $1e-5$ between the perplexity and the number of articles in each language in our dataset; we find a Spearman's $\rho$ of $-0.669$ and a p-value of $2e-4$ between the perplexity and the number of tokens in each language in the mC4 corpus where mT5 is pretrained on. These statistics suggest a significant correlation between perplexity and data size, while other typological factors are open to investigation.

Table~\ref{table:generation_ppl} also shows the BERTScore F1 measure of the generated scripts compared against the gold scripts. Except for English (.82), the performance across different languages varies within a relatively small margin (.65 - .72). However, we notice that as a metric based on the token-level pairwise similarity, BERTScore may not be the most suitable metric to evaluate scripts. It is best designed for \textit{aligned} texts (e.g. a machine-translated sentence and a human-translated one), whereas in scripts, certain candidate steps might not have aligned reference steps. Moreover, BERTScore does not measure whether the ordering among steps is correct. To address these flaws, we further perform human evaluation in Section~\ref{section:human_eval}.

\subsection{Auto Evaluation for Retrieval Setting}
\label{section:retrieval_results}
To automatically evaluate models in the Retrieval Setting, we first calculate \textbf{accuracy}, i.e. the percentage of predicted steps that exist in the ground-truth steps. To account for the ordering of steps, we also compute \textbf{Kendall's $\tau$} between the overlapping steps in the prediction and the ground-truth.


The performance of our Step Inference-Ordering pipeline using mBERT and XLM-RoBERTa\footnote{XLM-RoBERTa is not able to converge on the training data for Step Ordering for all but 3 languages using a large set of hyperparameter combinations.} on all 18 languages are shown in Figure~\ref{fig:charts}. Complete results can be found in Appendix~\ref{appendix:results}. Across languages, the results are generally similar with a large room for improvement. On average, our best system constructs scripts with around $30\%$ accuracy and around $0.2$ Kendall's $\tau$ compared to the ground-truth. Compared to the baseline, our multitask and cross-lingual zero-shot variations demonstrate significant improvement on ordering. This is especially notable in low-resource languages. For example, MTL on Korean and C0 on Thai both outperform their baseline by 0.17 on Kendall's $\tau$.


\subsection{Human Evaluation}
\label{section:human_eval}

To complement automatic evaluation, we ask 6 annotators\footnote{The annotators are graduate students and native or proficient speakers of the language assigned.} to each \textit{edit} 30 output scripts by the Step-Inference-Ordering pipeline and mT5 in English, French, Chinese, Japanese, Korean and Hindi, respectively. The \textit{edit} process consists of a sequence of two possible actions: either 1) delete a generated step entirely if it is irrelevant, nonsensical or not a reasonable step of the given goal, or 2) move a step somewhere else, if the order is incorrect. Then, the generated script is evaluated against the edited script in 3 aspects:\\
\textbf{Correctness}, approximated by the length (number of steps) of the edited script over that of the originally constructed script (c.f. precision);\\
\textbf{Completeness}, approximated by the length of the edited script over that of the ground-truth script (c.f. recall);\\
\textbf{Orderliness}, approximated by Kendall's $\tau$ between overlapping steps in the edited script and the generated script.\footnote{In this formulation, the correctness and completeness of a retrieval-based model are equal, since the length of its constructed script is equal to that of the ground truth script by definition.}

The results are shown in Table~\ref{table:generation_human_results}. While the constructed scripts in the Retrieval setting contain more correct steps, their ordering is significantly worse than those in the Generation setting. This suggests that the generation model is better at producing \textit{fluent} texts, but can easily suffer from hallucination.

\begin{table}[t!]
\small
\centering
\begin{tabular}{ l|cccccc } 
\toprule
\multicolumn{7}{c}{Retrieval: Step-Inference-Ordering pipeline} \\
\midrule
Language & en & fr & zh & jp & ko & hi  \\
Correctness & .70 & .39 & .50 & .49 & .45 & .82  \\
Completeness & .70 & .39 & .50 & .49 & .45 & .82  \\
Orderliness & .45 & .38 & .16 & .12 & .10 & .75  \\
\bottomrule
\toprule
\multicolumn{7}{c}{Generation: mT5} \\
\midrule
Language & en & fr & zh & jp & ko & hi  \\
Correctness & .39 & .51 & .46 & .40 & .37 & .49  \\
Completeness & .35 & .40 & .46 & .30 & .36 & .41  \\
Orderliness & .82 & .46 & .60 & .81 & .69 & .88  \\
\bottomrule
\end{tabular}
\caption{Human judgments of correctness, completeness and orderliness of the output of the Step-Inference-Order pipeline and the mT5 model for the same set of 30 gold scripts, in six languages. }
\label{table:generation_human_results}
\end{table}

\begin{figure*}[t!]
    \centering
    \includegraphics[scale=0.19]{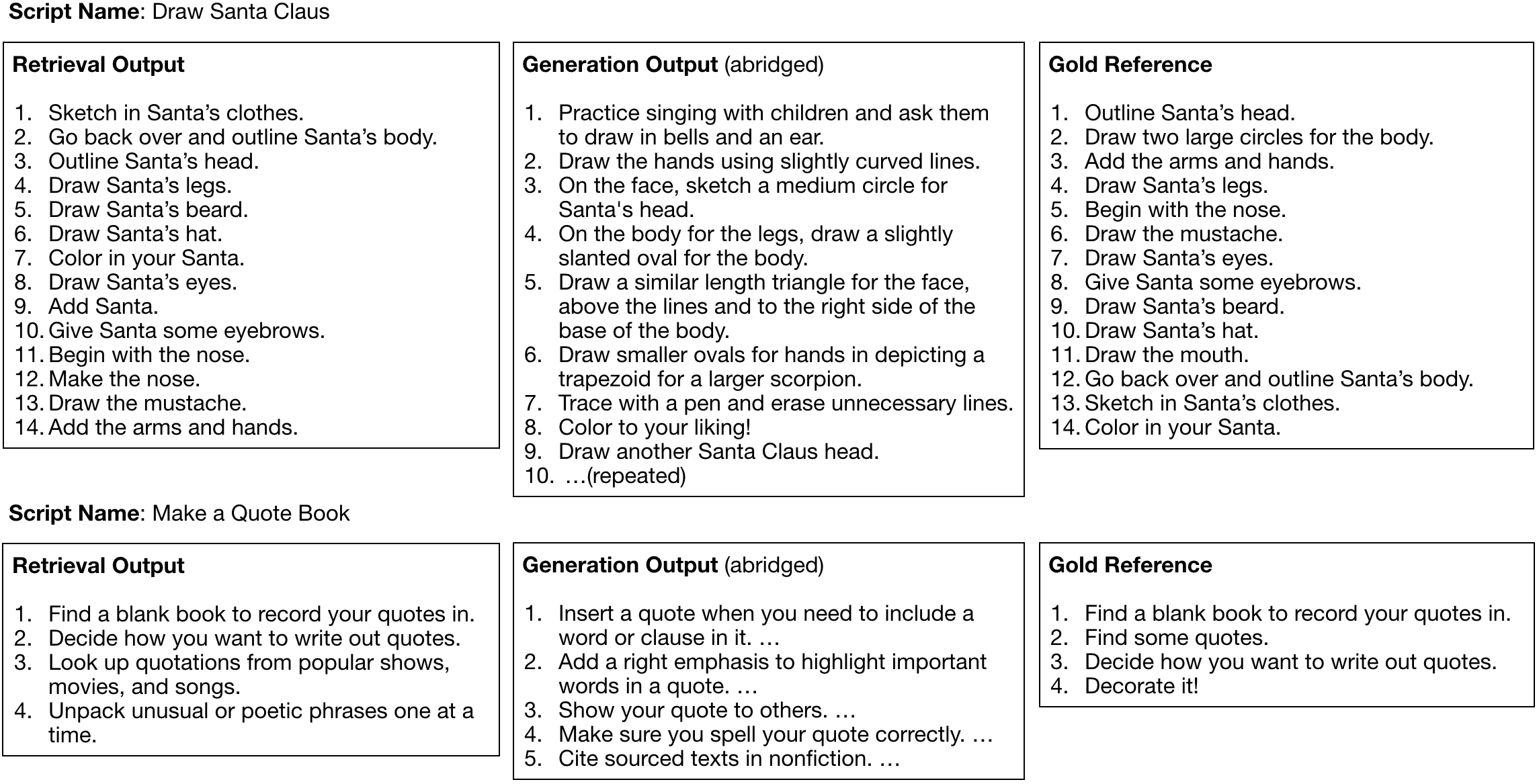}
    \caption{Two example scripts constructed by our Retrieval and Generation approaches.}
    \label{fig:qual_example}
\end{figure*}

\subsection{Qualitative Examples}
\label{section:in_domain_qual_examples}
To understand models' behavior, we present two representative scripts produced by the mBERT Retrieval model and the mT5 Generation model side by side, accompanied by the ground-truth script, shown in Figure~\ref{fig:qual_example}.

The retrieved ``Draw Santa Claus'' script has a high step accuracy (85\%), with a reasonable ordering of drawing first the outline and then details. The generation output is more off-track, hallucinating irrelevant details like ``singing'' and ``scorpion'', despite being on the general topic of drawing. It also generates more repetitive steps (e.g. the head is drawn twice), most of which are abridged. 

As for ``Make a Quotebook'', the retrieved script has a 50\% step accuracy. The third step, though not in the gold reference, is similar enough to ``find some quotes'', suggesting that our exact match evaluation isn't perfect. In the generated script, all steps are also generally plausible, but some essential steps are missing (e.g. find a book, find quotes). This suggests that the generation model dwells too much on the details, ignoring the big picture.

These patterns in the two scripts are common in the model outputs, a larger sample of which is included in the Supplementary Materials.

\section{Zero-shot Transfer Learning}

To show the potential of our model for transfer learning, we use the retrieval-based Step-Inference-Ordering pipeline finetuned on wikiHow to construct scripts for other datasets and domains. We quantitatively evaluate our model on 4 other script learning corpora, and qualitatively analyze some constructed scripts in a case study.


\subsection{Quantitative Evaluation}
Since no multilingual script data are available yet, we perform transfer learning experiments on 4 other English script corpora, OMICS \cite{singh2002open}, SMILE \cite{regneri2010learning}, DeScript  \cite{wanzare2016crowdsourced} \footnote{The above 3 corpora are all obtained from \url{http://www.coli.uni-saarland.de/projects/smile/}}, and the KAIROS Schema Learning Corpus (LDC2020E25). The first 3 pertain to human activities, while the last is in the military and political domain. They are all in the format of different \textit{scenarios} (e.g. ``eat in a restaurant'', similar to our \textit{goal}) each with a number of \textit{event sequence descriptions} (ESDs, similar to our \textit{steps}). Statistics for each corpus are in Table~\ref{table:transfer_results}.

\begin{table}
\small
\centering
\begin{tabular}{ l|cc|cccc|c|cc } 
 \toprule
\multirow{2}{*}{\begin{tabular}[c]{@{}l@{}}Corpus \end{tabular}} & \multicolumn{2}{c|}{Corpus Stats.}  & \multicolumn{2}{c}{Results} \\
& Scenarios & ESDs  & Acc.   & Kendall's $\tau$   \\
 \midrule
SMILE & 22 & 386 & .435 & .391\\
OMICS  & 175  & 9044 & .346 & .443\\
DeScript & 40  & 4000 & .414 & .418\\
KAIROS  & 28  & 28 & .589 & .381\\
\bottomrule
\end{tabular}
\caption{The zero-shot GOSC Retrieval performance of XLM-RoBERTa finetuned on wikiHow on 4 target corpora.}
\label{table:transfer_results}
\end{table}

For each dataset, we select the ESD with the most steps for every scenario as a representative script to avoid duplication,
thus converting the dataset to a GOSC evaluation set under the Retrieval setting. We then use the XLM-RoBERTa-based Step-Inference-Ordering pipeline trained on our English wikiHow dataset to directly construct scripts on each target set, and report its zero-shot performance in Table~\ref{table:transfer_results}. We see that $30\%-60\%$ steps are accurately retrieved, and around $40\%$ are correctly ordered. This is close to or even better than the in-domain results on our English test set. As a comparison, a random baseline would have only 0.013 Accuracy and 0.004 $\tau$ on average. Both facts indicate that the script knowledge learned from our dataset is clearly non-trivial.

\subsection{Case Study: The \textit{Bombing Attack} Scripts}
\label{section:case_study}
To explore if the knowledge about \textit{procedural} scripts learned from our data can also facilitate the zero-shot learning of \textit{narrative} scripts, we present a case study in the context of the DARPA KAIROS program\footnote{\href{https://tinyurl.com/yxwztj3j}{{\tt www.darpa.mil/program/knowledge- directed-artificial-intelligence-reasoning -over-schemas}}}. One objective of KAIROS is to automatically induce scripts from large-scale narrative texts, especially in the military and political domain. We show that models trained on our data of commonplace events can effectively transfer to vastly different domains.

With the retrieval-based script construction model finetuned on wikiHow, we construct five scripts with different granularity levels under the \textit{Improvised Explosive Device (IED) attack} scenario: ``Roadside IED attack'', ``Backpack IED attack'', ``Drone-brone IED attack'', ``Car bombing IED attack'', ``IED attack''. 
We take the name of each script as the input \textit{goal}, and a collection of related documents retrieved from Wikipedia and Voice of America news as data sources for extracting step candidates. 

Our script construction approach has two components. First, we extract all events according to the KAIROS Event Ontology from the documents using OneIE \cite{lin2020joint}. The ontology defines 68 event primitives, each represented by an \textit{event type} and multiple \textit{argument types}, e.g. a Damage-type event has arguments including Damager, Artifact, Place, etc. OneIE extracts all event instances of the predefined primitives from our source documents. Each event instance contains a \textit{trigger} and several \textit{arguments} (e.g. Trigger: ``destroy'', Damager: ``a bomber'', Artifact: ``the building'', ... ). All event instances form the candidate pool of steps for our target script.

Since the events are represented as trigger-arguments tuples, a conversion to the raw textual form is needed before inputting them into our model. This is done by automatically instantiating the corresponding event type template in the ontology with the extracted arguments. If an argument is present in the extracted instance, we directly fill it in the template; else, we fill in a placeholder word (e.g.``some'', ``someone'', depending on the argument type). For example, the template of Damage-type events is ``$\langle arg1 \rangle$ damaged $\langle arg2 \rangle$ using $\langle arg3 \rangle$ instrument'', which can be instantiated as ``A bomber damaged the building using some instrument''). Next, we run the Step Inference-Ordering Pipeline in Section~\ref{section:step_inf_order_pipeline} on the candidate pool given the ``goal''. The only modification is that since we don't have a gold reference script length in this case, all retrieved steps with a confidence score higher than a threshold (default=0.95) are retained in the final script.

\begin{figure}[t!]
    \centering
    \includegraphics[scale=0.15]{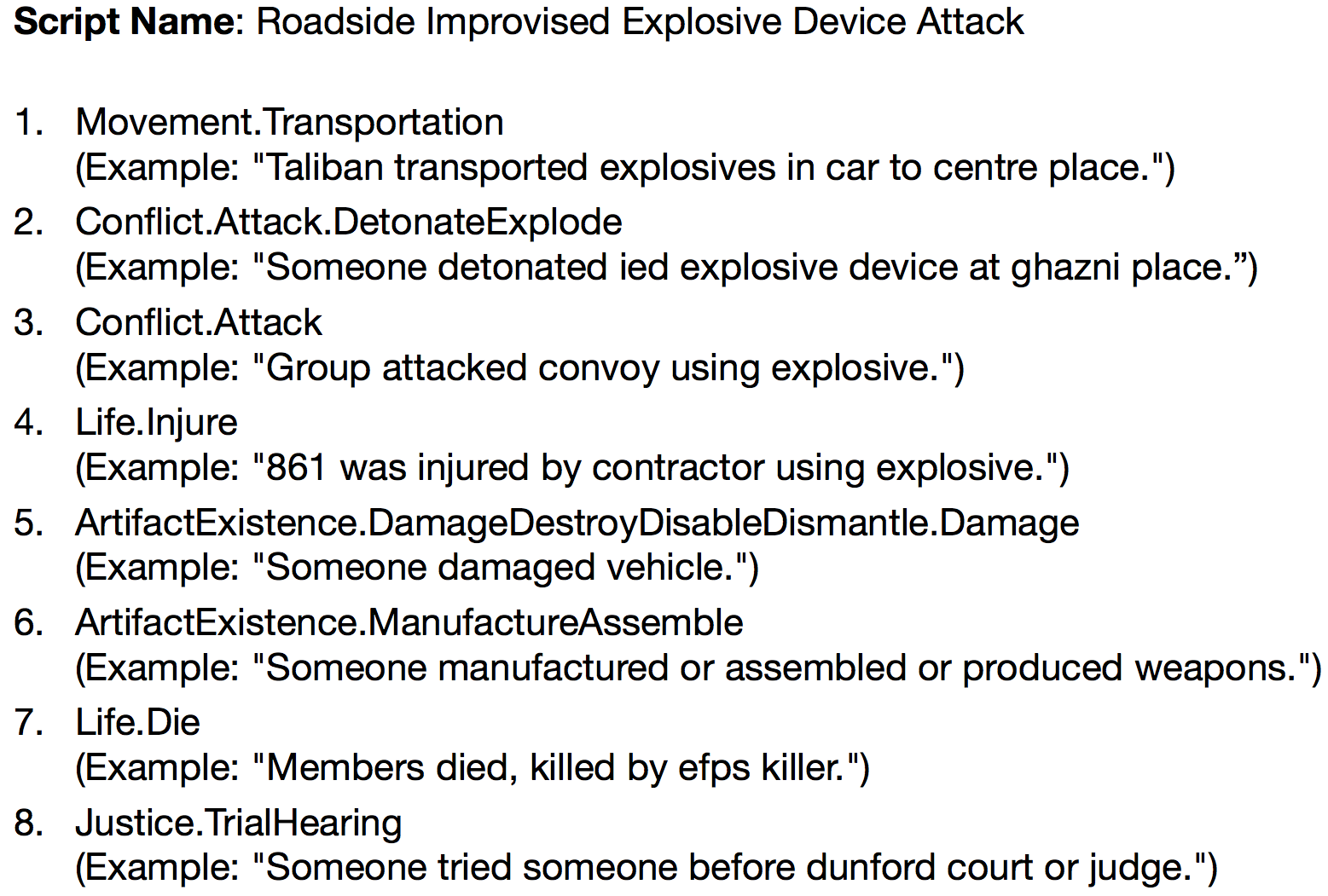}
    \caption{An example narrative script produced by our retrieval-based pipeline trained on wikiHow. Each event is represented by its Event Type and an example sentence.}
    \label{fig:IED_example}
\end{figure}

We manually evaluate the constructed scripts with the metrics defined in Section~\ref{section:human_eval}, except \textit{Completeness} as we don't have gold references. The 5 constructed scripts have an average \textit{Correctness} of 0.735 and \textit{Orderliness} of 0.404. Despite the drastic domain shift from wikiHow to KAIROS, our model can still exploit its script knowledge to construct scripts decently. An example script, ``Roadside IED attack'', is shown in Figure~\ref{fig:IED_example}. All the steps retrieved are sensible, and most are ordered with a few exceptions (e.g. the ManufactureAssemble event should precede all others).\footnote{More details on the format of the script, all five constructed scripts, the event ontology, and a list of news documents used can be found in the Supplementary Materials.} 

\section{Limitations}

\textbf{Event representation:} Our representation of goals and steps as natural language sentences, though containing richer information, brings the extra difficulty in handling steps with similar meanings. For example, ``change strings frequently'' and ``put on new strings regularly'' have nearly identical meanings and both are correct steps for the goal ``maintain a guitar''. Hence, both could be included by a retrieval-based model, which is not desired.\\
\textbf{Modeling:} Since GOSC is a new task, there is no previously established SOTA to compare with. We build a strong baseline for each setting, but they are clearly not the necessary or sufficient means to do the task. For example, our Step-Inference-Ordering pipeline would benefit from a paraphrasing module that eliminates semantic duplicates in retrieved steps. It also currently suffers from long run-time especially with a large pool of candidates, since it requires pairwise goal-step inference. An alternative is to filter out most irrelevant steps using similarity-based heuristics in advance.\\
\textbf{Evaluation:} Under the retrieval-based setting, our automatic evaluation metrics do not give credit to inexact matches as discussed above, which can also be addressed by a paraphrasing module. Meanwhile, for the generation-based setting, BERTScore, or other comparison-based metrics like BLEU \cite{papineni-etal-2002-bleu} and METEOR \cite{denkowski-lavie-2014-meteor}, may not be the most suitable metric to evaluate scripts. They are best designed for \textit{aligned} texts like translation pairs, and do not measure whether the ordering among steps is correct. While we complement it with manual evaluation, only one human annotator is recruited for each language, resulting in potential subjectivity. Alternatively, crowdsourcing-based evaluation is costly and hard to standardize. Due to the complexity of the GOSC task and its evaluation, we suggest that future work investigate better means of evaluation.

\section{Conclusion and Future Work}
We propose the first multilingual script learning dataset and the first task to evaluate the holistic understanding of scripts. By comprehensively evaluating model performances automatically and manually, we show that state-of-the-art models can produce complete scripts both in- and out-of-domain, with a large room for improvement. Future work should investigate additional aspects of scripts, such as usefulness, granularity, etc., as well as their utility for downstream tasks that require automated reasoning.

\section*{Acknowledgments}
This research is based upon work supported in part by the DARPA KAIROS Program (contract FA8750-19-2-1004), the DARPA LwLL Program (contract FA8750-19-2-0201), and the IARPA BETTER Program (contract 2019-19051600004). Approved for Public Release, Distribution Unlimited. The views and conclusions contained herein are those of the authors and should not be interpreted as necessarily representing the official policies, either expressed or implied, of DARPA, IARPA, or the U.S. Government.

Special thanks go to our annotators: Artemis Panagopoulou, Daniel Joongwon Kim, Simmi Mourya, and Liam Dugan. We also thank Daphne Ippolito for the help on mT5; Ying Lin, Sha Li, Zhenhailong Wang, Pengfei Yu, Tuan Lai, Haoyang Wen, and Heng Ji for providing the source documents and the OneIE results for our case study. Finally, we appreciate the valuable feedback from the anonymous reviewers.

\bibliography{acl2021}
\bibliographystyle{acl_natbib}

\vspace{0.3in}
\appendix

\section{Corpus Statistics}
\label{appendix:corpus_stats}

Table~\ref{table:corpus_stats} shows the statistics of our multilingual wikiHow script corpus.

\begin{table*}[t!]
\small
\centering
\begin{tabular}{ lccccccccc } 
 \toprule
Language & en & es & pt & de & fr & ru & it & id & zh \\
 \midrule
Num. articles & 112,111 & 64,725 & 34,194 & 31,541 & 26,309 & 26,222 & 22,553 & 21,014 & 14,725 \\
Num. ordered articles & 54,852 & 26,620 & 7,408 & 10,681 & 6,834 & 5,335 & 4,308 & 5,536 & 4,784 \\
Avg. num. of sections / article & 2.5 & 2.6 & 3.1 & 2.7 & 2.9 & 3.1 & 3.1 & 2.9 & 2.8 \\
Avg. num. of steps / article & 13.7 & 14.4 & 16.1 & 15.1 & 15.7 & 16.4 & 16.2 & 16.0 & 15.4 \\
Num. of articles for train/dev & 100,900 & 58,253 & 30,775 & 28,387 & 23,679 & 23,600 & 20,298 & 18,913 & 13,253 \\
Num. of articles for test & 11,211 & 6,472 & 3,419 & 3,154 & 2,630 & 2,622 & 2,255 & 2,101 & 1,472 \\
 \bottomrule
 \toprule
Language & nl & ar & vn & th & jp & ko & cz & hi & tr \\
 \midrule
Num. articles & 13,343 & 12,157 & 6,949 & 5,821 & 5,567 & 5,179 & 5,043 & 3,104 & 1,434 \\
Num. ordered articles & 2,113 & 2,567 & 1,157 & 1,244 & 1,209 & 917 & 920 & 888 & 520 \\
Avg. num. of sections / article & 3.1 & 3.0 & 3.2 & 3.1 & 3.1 & 3.2 & 3.1 & 3.0 & 3.2 \\
Avg. num. of steps / article & 16.2 & 16.4 & 17.1 & 17.7 & 16.8 & 17.5 & 16.4 & 16.8 & 19.2 \\
Num. of articles for train/dev & 12,009 & 10,942 & 6,255 & 5,239 & 5,011 & 4,662 & 4,539 & 2,794 & 1,291 \\
Num. of articles for test & 1,334 & 1,215 & 694 & 582 & 556 & 517 & 504 & 310 & 143 \\

 \bottomrule
\end{tabular}
\caption{Statistics of our multilingual wikiHow corpus by language, ordered by the number of articles in each language. Each article is converted to a script, including all steps from all sections.}
\label{table:corpus_stats}
\end{table*}

\section{Evaluation Details}
\label{appendix:metrics}
In Section~\ref{section:task}, we formalize the Goal-Oriented Script Construction (GOSC) task as follows: Given a \textit{goal} $g$, the model is asked to construct a complete script as an ordered list of \textit{steps} $S$, with a ground-truth reference $T$. As a hint of the desired level of granularity, we also provide an expected number of steps (or length of the script), $l$, as input.

In the \textbf{Retrieval setting}, a set of candidate steps $C$ is also available. We evaluate an output script from two angles: content and ordering. 

First, we calculate the accuracy, namely the percentage of predicted steps that exist in the ground-truth. Denote $s_i$ as the $i$-th step in $S$. 
\[ \textrm{acc} = (\sum_i^l [s_i \in T])/l \]
If the gold script is ordered, we further evaluate the ordering of the constructed script by calculating Kendall's $\tau$ between the intersection of the predicted steps and the ground-truth steps.
\[\tau = \frac{NC(S\cap T,T\cap S)-ND(S\cap T,T\cap S)}{\binom{l}{2}}\]
where $NC$ is the number of concordant pairs, $ND$ the number of discordant pairs; $A \cap B$ is used as a special notation for the intersection of ordered lists, denoting elements that appear in both $A$ and $B$, in the order of $A$. 

It is likely that a model includes two modules: a retrieval module and an ordering module. In this case, it is sensible to separately evaluate these two modules. 

To evaluate the retrieval module independently, assume that the model retrieves a large set of steps $R$ ranked by their \textit{relevance} to the goal $g$. Denote $r_i$ as the $i$-th step in $R$.  We calculate recall and normalized discounted cumulative gain\footnote{We set the true relevance of each predicted step as 1 if it exists in the ground-truth steps, and 0 otherwise.} at position $k$. Assume $k>l$.
\[recall_k = (\sum_i^k [r_i \in T])/k\]
\[NDCG_k = \frac{\sum_{i=1}^k \frac{2^{[r_i \in T]}-1}{\log_2(i+1)}}{\sum_{i=1}^l \frac{2^{1}-1}{\log_2(i+1)} + \sum_{i=l+1}^k \frac{2^{0}-1}{\log_2(i+1)}}\]

To evaluate the ordering module independently, we directly give the model the set of ground-truth steps to predict an ordering. We again use Kendall's $\tau$ to evaluate the ordered steps.
\[\tau = \frac{NC(T',T)-ND(T',T)}{\binom{l}{2}}\]
where $T'$ is the set ground-truth steps ordered by the model.

In the \textbf{Generation} setting, a model is evaluated using perplexity on the test set, following standard practice.
\[perplexity(S)=\exp{(-\frac{L(S)}{\text{count of tokens in } S})}\]
where $L(S)$ is the log-likelihood of the sequence of steps assigned by the model.

When evaluating a model on multiple scripts, all aforementioned metrics are averaged. 

\section{Modeling Details}
\label{appendix:model}
All our models are implemented using the HuggingFace Transformer service\footnote{\url{https://github.com/huggingface/transformers}}. For all experiments, we hold out 5\% of the training data for development. 

The pretrained models we use include: the {\small\verb|bert-base-multilingual-uncased|} checkpoint (168M parameters) for mBERT, the {\small\verb|xlm-roberta-base|} checkpoint (270M parameters) for XLM-RoBERTa, the {\small\verb|roberta-base|} checkpoint (125M parameters) for RoBERTa\footnote{The above 3 models are available at \url{https://huggingface.co/transformers/pretrained_models.html}}, and the {\small\verb|mT5-Large|} checkpoint (1B parameters) for mT5\footnote{\url{https://github.com/google-research/multilingual-t5}}. 

For mBERT, XLM-RoBERTa and RoBERTa, we finetune the pretrained models on our dataset using the standard {\small\verb|SequenceClassification|} pipeline on HuggingFace\footnote{\url{https://huggingface.co/transformers/model_doc/auto.html?highlight=sequence\%20classification\#transformers.AutoModelForSequenceClassification}}. For mT5, we refer to the official finetuning scripts\footnote{\url{https://colab.research.google.com/github/google-research/text-to-text-transfer-transformer/blob/master/notebooks/t5-trivia.ipynb}} from the project's Github repository.

For each in-domain evaluation experiment, we perform grid search on learning rate from $1e-5$ to $5e-8$, batch size from $16$ to $128$ whenever possible, and the number of epochs from $3$ to $10$. As mBERT and XLM-RoBERTa have a large number of hyperparameters, most of which remain default, we do not list them here. Instead, the hyperparameter values and pretrained models will be available publicly via HuggingFace model sharing. We choose the model with the highest validation performance to be evaluated on the test set. For the Retrieval setting, we consider the accuracy of contracted scripts; for the Generation setting, we consider perplexity. 

We run our experiments on an NVIDIA GeForce RTX 2080 Ti GPU, with half-precision floating point format (FP16) with O1 optimization. The experiments in the Retrieval setting take 3 hours to 5 days in the worst case for all languages. The experiments in the Generation setting take 2 hours to 1 day in the worst case for all languages. 

\section{Additional Results}
\label{appendix:results}
Our complete in-domain evaluation results can be found in Table~\ref{table:mtl_results}, \ref{table:results}, and~\ref{table:c0_results}.

\begin{table}[t!]
\small
\centering
\begin{tabular}{ l|ccccccccc } 
\toprule
Lang. & en & es & pt & de & fr & ru  \\
\midrule
Acc. & .253 & .220 & .248 & .266 & .248 & \textbf{.275}  \\
$\tau$ & \textbf{.371} & \textbf{.313} & \textbf{.225} & \textbf{.249} & \textbf{.269} & .244  \\
\midrule
\midrule
Lang. & it & id & zh & nl & ar & vn \\
\midrule
Acc. & .261 & \textbf{.293} & \textbf{.322} & \textbf{.288} & .276 & .311  \\
$\tau$ & .208 & \textbf{.195} & \textbf{.237} & \textbf{.184} & .164 & .183  \\
\midrule
\midrule
Lang. & th & jp & ko & cz & hi & tr \\
\midrule
Acc. & .100 & \textbf{.325} & \textbf{.325} & .310 & .349 & .415  \\
$\tau$ & .124 & \textbf{.220} & \textbf{.227} & \textbf{.293} & \textbf{.215} & .148  \\
\bottomrule
\end{tabular}
\caption{The GOSC Retrieval performance of multitask learning mBERT. Results higher than those produced by the single-task mBERT are in bold.}
\label{table:mtl_results}
\end{table}

\begin{table*}
\centering
\begin{tabular}{ l|cccc|c|cc } 
 \toprule
\multirow{2}{*}{\begin{tabular}[c]{@{}l@{}}Lang. \end{tabular}} & \multicolumn{4}{c|}{Step Retrieval}                  & \multicolumn{1}{c|}{Ordering} & \multicolumn{2}{c}{Script Construction} \\
& Recall@25 & Recall@50 & NDCG@25 & NDCG@50 & Kendall's $\tau$                & Accuracy   & Kendall's $\tau$   \\
 \midrule
en & .337 / .342 & .424 / .429 & .660 / .660 & .648 / .648 & .368 / .375 & .256 / .258 & .369 / .372 \\
es & .319 / .397 & .403 / .786 & .653 / .532 & .642 / .571 & .321 / .022 & .246 / .216 & .295 / .022 \\
pt & .313 / .319 & .401 / .412 & .679 / .672 & .664 / .659 & .207 / .212 & .251 / .254 & .186 / .202 \\
de & .337 / .350 & .421 / .438 & .687 / .707 & .676 / .692 & .260 / .026 & .268 / .280 & .276 / .048 \\
fr & .315 / .320 & .405 / .411 & .673 / .672 & .661 / .659 & .244 / .020 & .248 / .250 & .206 / .043 \\
ru & .336 / .353 & .423 / .446 & .701 / .715 & .688 / .701 & .181 / .042 & .271 / .285 & .207 / .006 \\
it & .332 / .333 & .424 / .431 & .700 / .705 & .686 / .687 & .184 / .035 & .264 / .267 & .218 / .081 \\
id & .351 / .383 & .435 / .480 & .712 / .744 & .699 / .725 & .190 / .011 & .284 / .312 & .182 / .026 \\
zh & .401 / .429 & .498 / .536 & .750 / .753 & .732 / .737 & .260 / .027 & .319 / .340 & .203 / .030 \\
nl & .354 / .382 & .447 / .758 & .721 / .546 & .708 / .597 & .179 / .011 & .287 / .243 & .156 / .075 \\
ar & .351 / .381 & .447 / .485 & .710 / .735 & .694 / .717 & .161 / .055 & .288 / .311 & .175 / .067 \\
vn & .381 / .436 & .464 / .544 & .769 / .784 & .753 / .766 & .170 / .171 & .322 / .358 & .206 / .152 \\
th & .146 / .448 & .273 / .566 & .330 / .784 & .369 / .764 & .106 / .056 & .104 / .362 & .149 / .048 \\
jp & .383 / .447 & .487 / .579 & .754 / .766 & .732 / .751 & .170 / .107 & .308 / .356 & .179 / .097 \\
ko & .381 / .435 & .474 / .553 & .762 / .780 & .744 / .766 & .154 / .044 & .318 / .361 & .056 / .068 \\
cz & .416 / .456 & .532 / .582 & .772 / .776 & .751 / .758 & .211 / -.007 & .337 / .367 & .190 / -.028 \\
hi & .421 / .484 & .530 / .610 & .782 / .814 & .763 / .798 & .156 / .029 & .354 / .410 & .138 / -.004 \\
tr & .509 / .577 & .676 / .718 & .859 / .881 & .829 / .854 & .154 / .014 & .415 / .477 & .176 / .047 \\
\midrule
Mean & .355 / .404 & .454 / .542 & .704 / .724 & .691 / .714 & .204 / .069 & .286 / .317 & .198 / .075 \\
\bottomrule
\end{tabular}
\caption{The GOSC Retrieval performance of mBERT and XLM-RoBERTa, divided by a slash in each cell. Both the performance of individual modules and that of script construction are reported.}
\label{table:results}
\end{table*}

\begin{table*}
\centering
\begin{tabular}{ l|cccc|c|cc } 
 \toprule
\multirow{2}{*}{\begin{tabular}[c]{@{}l@{}}Lang. \end{tabular}} & \multicolumn{4}{c|}{Step Retrieval}                  & \multicolumn{1}{c|}{Ordering} & \multicolumn{2}{c}{Script Construction} \\
& Recall@25 & Recall@50 & NDCG@25 & NDCG@50 & Kendall's $\tau$                & Accuracy   & Kendall's $\tau$   \\
 \midrule
es & 0.270 & 0.338 & 0.567 & 0.564 & 0.360 & 0.203 & 0.353\\
pt & 0.265 & 0.339 & 0.595 & 0.590 & 0.276 & 0.212 & 0.310\\
de & 0.271 & 0.346 & 0.556 & 0.558 & 0.264 & 0.206 & 0.285\\
fr & 0.253 & 0.319 & 0.585 & 0.580 & 0.283 & 0.202 & 0.268\\
ru & 0.313 & 0.390 & 0.672 & 0.661 & 0.252 & 0.258 & 0.284\\
it & 0.264 & 0.338 & 0.575 & 0.574 & 0.268 & 0.210 & 0.317\\
id & 0.338 & 0.424 & 0.681 & 0.670 & 0.321 & 0.277 & 0.293\\
zh & 0.379 & 0.471 & 0.718 & 0.706 & 0.318 & 0.310 & 0.295\\
nl & 0.340 & 0.424 & 0.684 & 0.673 & 0.280 & 0.280 & 0.263\\
ar & 0.319 & 0.400 & 0.643 & 0.635 & 0.235 & 0.260 & 0.211\\
vn & 0.392 & 0.480 & 0.748 & 0.733 & 0.249 & 0.333 & 0.255\\
th & 0.418 & 0.520 & 0.771 & 0.753 & 0.307 & 0.356 & 0.320\\
jp & 0.403 & 0.512 & 0.751 & 0.733 & 0.232 & 0.338 & 0.238\\
ko & 0.391 & 0.485 & 0.767 & 0.749 & 0.182 & 0.336 & 0.218\\
cz & 0.406 & 0.519 & 0.749 & 0.732 & 0.300 & 0.342 & 0.285\\
hi & 0.449 & 0.557 & 0.770 & 0.754 & 0.205 & 0.384 & 0.187\\
tr & 0.505 & 0.646 & 0.805 & 0.787 & 0.167 & 0.434 & 0.099\\
\midrule
Mean & 0.357 & 0.448 & 0.692 & 0.681 & 0.259 & 0.296 & 0.258\\
\bottomrule
\end{tabular}
\caption{The GOSC Retrieval performance of XLM-RoBERTa using cross-lingual zero-shot transfer trained on the English data. Both the performance of individual modules and that of script construction are reported.}
\label{table:c0_results}
\end{table*}

\section{More Qualitative Examples}
\label{appendix:examples}
\begin{figure*}[t!]
    \centering
    \includegraphics[scale=0.2]{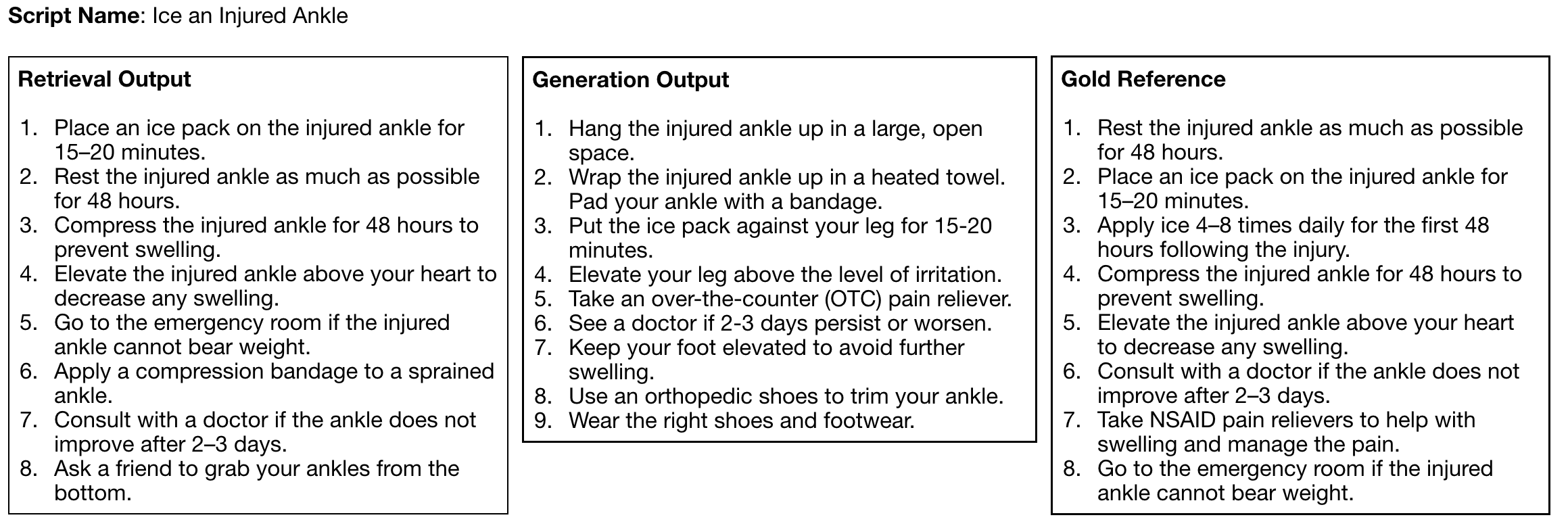}
    \caption{The ``Ice an injured Ankle'' script.}
    \label{fig:en1}
\end{figure*}

\begin{figure*}[t!]
    \centering
    \includegraphics[scale=0.2]{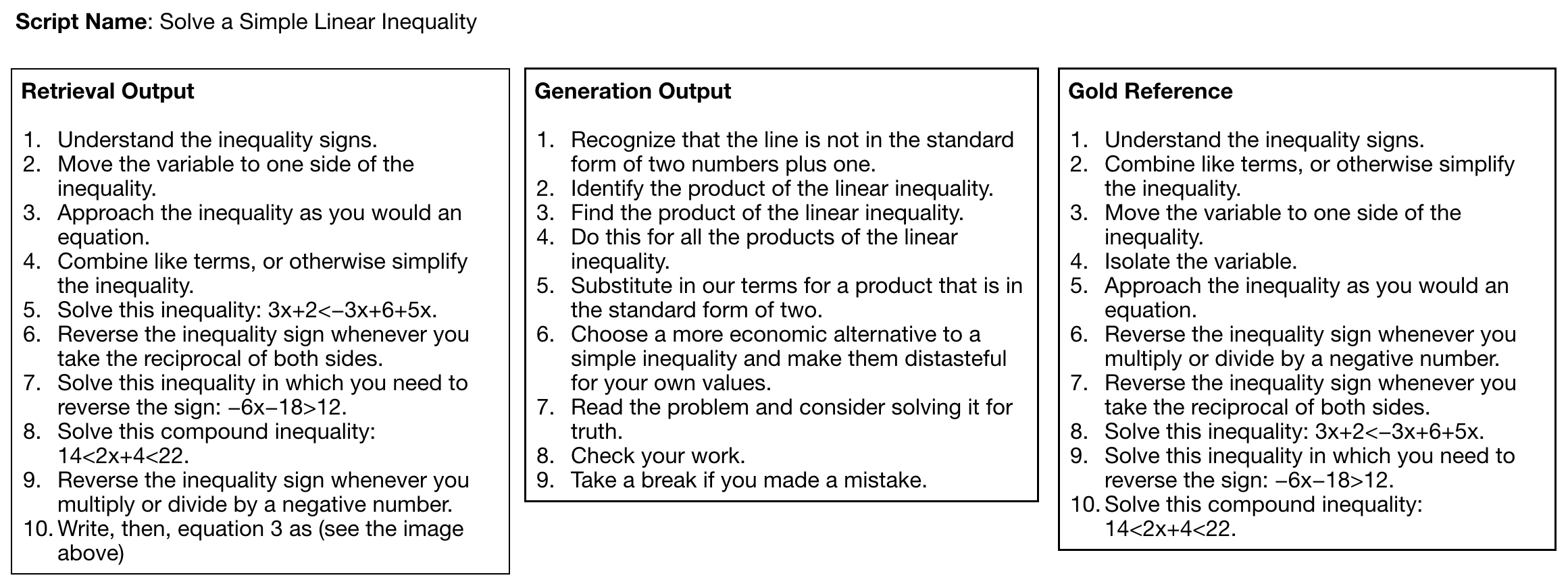}
    \caption{The ``Solve a Simple Linear Inequality'' script.}
    \label{fig:en2}
\end{figure*}

Aside from the examples shown in Section~\ref{section:in_domain_qual_examples}, we show 2 more example scripts constructed by the mBERT baseline under the Retrieval setting in Section~\ref{section:step_inf_order_pipeline} vs. those by the mT5 baseline under the Generation setting in Section~\ref{section:generation_baseline}. For each script name, the Retrieval output and the Generation output are shown side by side. Please see Figure~\ref{fig:en1} and ~\ref{fig:en2} for English examples, and Figure~\ref{fig:zh1} and \ref{fig:zh2} for Chinese ones.

For more examples, please see the Supplementary Materials. We include 20 examples for each language for the in-domain evaluation, and all 5 examples for the out-of-domain case study on the \textit{Bombing Attack} scenario.

    \begin{figure*}[t!]
        \centering
        \includegraphics[scale=0.1]{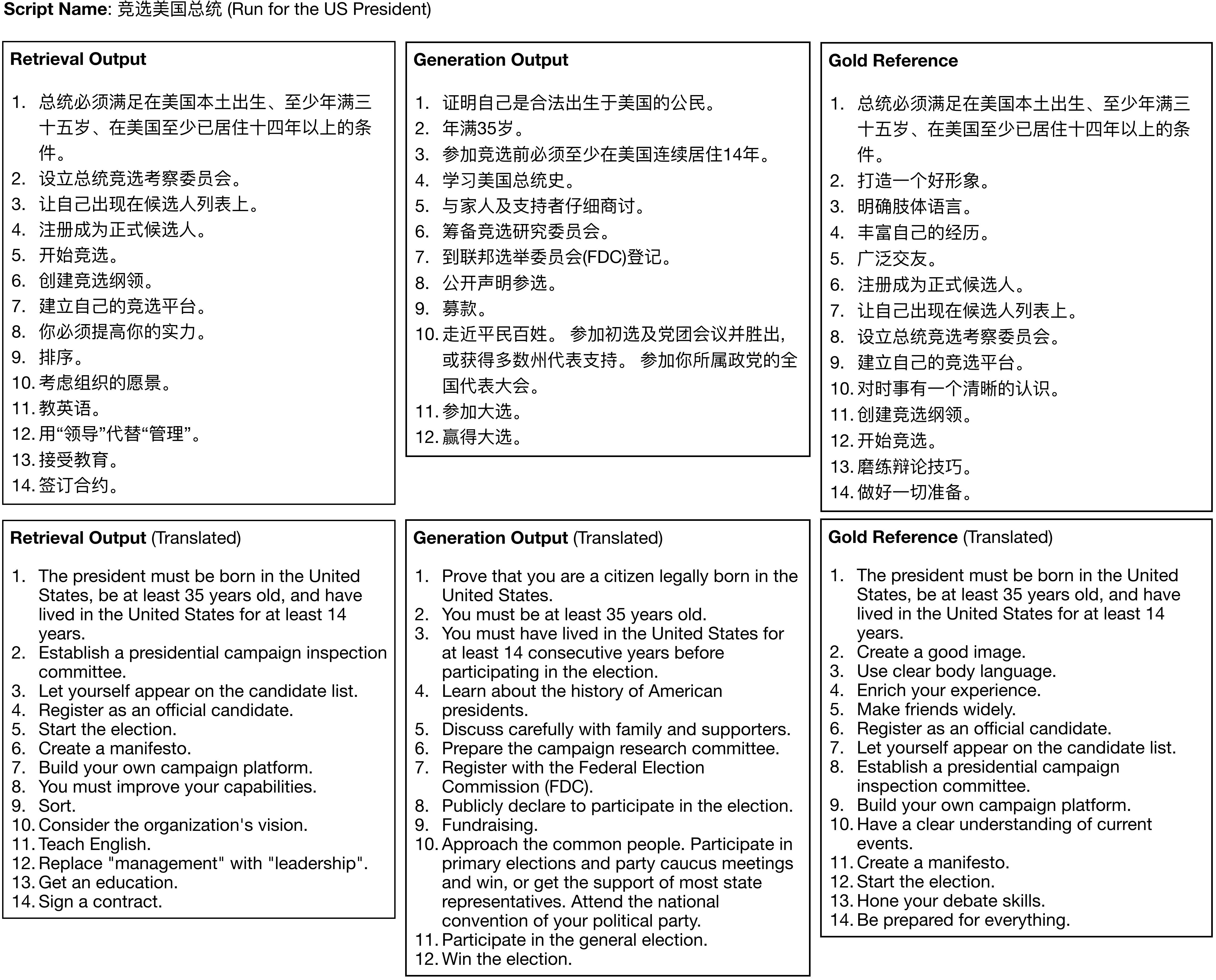}
        \caption{The ``Run for the US President'' script.}
        \label{fig:zh1}
    \end{figure*}
    \begin{figure*}[t!]
        \centering
        \includegraphics[scale=0.1]{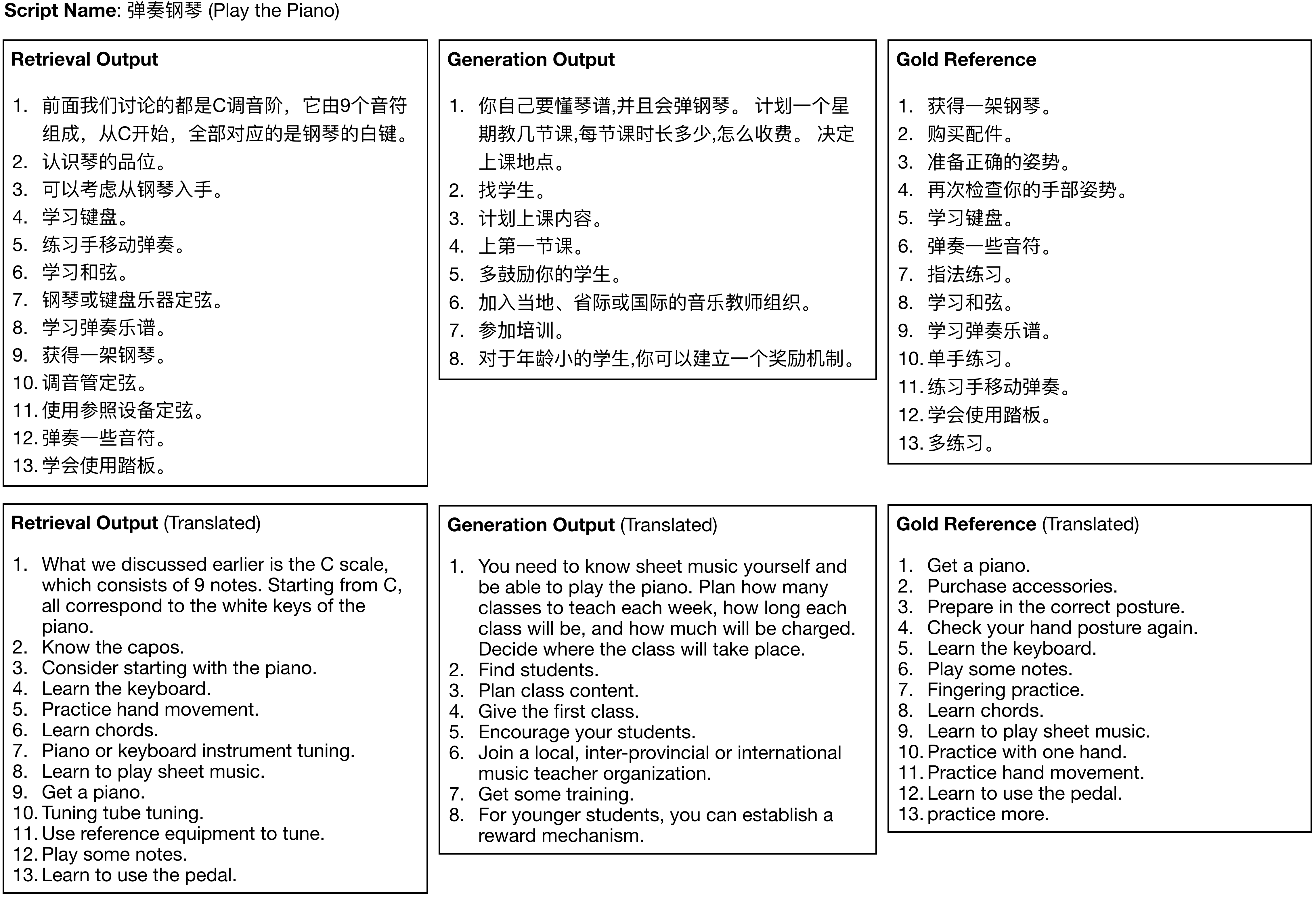}
        \caption{The ``Play the Piano'' script.}
        \label{fig:zh2}
    \end{figure*}

\end{document}